\pdfoutput=1

\documentclass[11pt]{article}

\usepackage[]{ACL2023}

\usepackage{times}
\usepackage{latexsym}
\usepackage{booktabs}
\usepackage{amssymb}

\usepackage[T1]{fontenc}

\usepackage[utf8]{inputenc}

\usepackage{microtype}

\usepackage{inconsolata}

\title{SynCPKL: Harnessing LLMs to Generate Synthetic Data for Commonsense Persona Knowledge Linking}

\author{Kuan-Yen Lin \\
  \texttt{iris19132@gmail.com}}

\begin{document}
\maketitle
\begin{abstract}
Understanding rich dialogues often requires NLP systems to access relevant commonsense persona knowledge, but retrieving this knowledge is challenging due to complex contexts and the implicit nature of commonsense. This paper presents our approach to the Commonsense Persona Knowledge Linking (CPKL) challenge, addressing the critical need for integrating persona and commonsense knowledge in open-domain dialogue systems. 
We introduce SynCPKL Pipeline, a pipeline that leverages Large Language Models to generate high-quality synthetic datasets for training commonsense persona knowledge linkers. 
To demonstrate the efficacy of our approach, we present SynCPKL, a new dataset specifically designed for this task. 
Our experiments validate the effectiveness of SynCPKL for training commonsense persona knowledge linkers. 
Additionally, our top-performing model, Derberta-SynCPKL, secured first place in the CPKL challenge by a 16\% improvement in F1 score.
We released both SynCPKL and Derberta-SynCPKL at \url{https://github.com/irislin1006/CPKL}.
\end{abstract}

\section{Introduction}
The field of human-computer interaction has seen steady progress (\citealp[]{Lee2022CoAuthorDA, Zhou2023SOTOPIAIE, Han2023RECIPEHT}), particularly in open-domain dialogue systems (\citealp[]{Jang2023ConversationCT, Zhang2023MindTG, Bae2022BuildingAR, Han2022MeetYF}). While Large Language Models (LLMs) have significantly improved the human-like quality of conversational agents (\citealp[]{Achiam2023GPT4TR, Maharana2024EvaluatingVL}), challenges persist in maintaining long-term memory, consistent persona attributes, and rich dialogue engagement. To address these issues, researchers are exploring the integration of persona and commonsense knowledge into dialogue systems (\citealp[]{Jandaghi2023FaithfulPC, Lee2022PERSONACHATGENGP}). Notable efforts include the ComFact \citep{Gao2022ComFactAB} benchmark for identifying situationally relevant commonsense facts, and PeaCok \citep{Gao2023PeaCoKPC}, a world-level persona commonsense knowledge graph designed to enhance open-domain conversations.

The extraction of pertinent persona-based commonsense information from existing knowledge bases presents significant challenges, stemming from the intricate and multifaceted nature of real-world conversations. This complexity is further compounded by the inherent subtlety and frequent ambiguity of commonsense knowledge itself. The nuanced interplay between dialogue context and persona-specific information often eludes traditional retrieval methods, highlighting the need for more sophisticated approaches. 

To address these challenges, we present our innovative solution to the Commonsense Persona Knowledge Linking (CPKL) challenge (\citealp[]{Wakaki2024ComperDialCP}). This shared task calls for robust commonsense persona knowledge linkers capable of identifying and seamlessly integrating relevant commonsense facts associated with both speakers and listeners in dialogues. In addressing this challenge, we face a fundamental obstacle: the lack of high-quality annotated datasets for training and evaluating commonsense persona knowledge linkers. To overcome this problem, we propose leveraging the grokking capabilities of LLMs to generate synthetic datasets that capture the complexities of commonsense persona knowledge in dialogues. Our method, SynCPKL Pipeline, aims to distill the implicit understanding of personal and social dynamics embedded in LLMs into explicit and structured datasets suitable for training commonsense persona knowledge linkers (\citealp[]{Lee2022DialogCCAA}). 
This approach not only addresses the scarcity of suitable training data but also allows for the creation of brand-new tasks lacking pre-built datasets.

Using the SynCPKL pipeline, we present SynCPKL, a new dataset specifically designed for training commonsense persona knowledge linkers. Through our experiments and analysis, we demonstrate the efficacy of SynCPKL in this domain. To foster further research and innovation, we have made SynCPKL publicly available to the research community.

Furthermore, we showcase Derberta-SynCPKL, our best-performing model that achieved first place in the CPKL challenge. By open-sourcing this model, we aim to accelerate progress in the field and provide a strong baseline for future research. Derberta-SynCPKL demonstrates the practical application of our synthetic data approach, highlighting its potential to drive significant improvements in commonsense persona knowledge linking tasks.

\section{Shared Task Setup}
The Commonsense Persona Knowledge Linking for Dialogue (CPKL) challenge from the 6th Workshop on NLP for ConvAI aims to develop models that link relevant commonsense persona knowledge to open-domain dialogues. This task is crucial for enhancing NLP systems' ability to ground rich dialogues in appropriate commonsense knowledge.

The challenge requires participants to create a model that determines whether a given persona commonsense fact is relevant to a speaker in a dialogue context. Each example consists of:
\begin{itemize}
    \item A dialogue between two speakers with a window size of 5 utterances ([\texttt{ut-2}, \texttt{ut-1}, \texttt{ut}, \texttt{ut+1}, \texttt{ut+2}]), where the target speaker is associated with utterance \texttt{ut}.
    \item A persona commonsense fact triple (head, relation, tail) from the PeaCoK knowledge graph.
\end{itemize}

The task does not provide a training dataset, allowing participants to utilize any datasets they deem appropriate. 
Evaluation is conducted on a closed test set using F1 score as the primary metric, with accuracy as a secondary measure.

\subsection{PeaCoK Knowledge Graph}
PeaCoK (Persona Commonsense Knowledge) is a large-scale knowledge graph designed to enhance dialogue systems' consistency and engagement through persona-based commonsense knowledge. 

In PeaCoK, a persona fact is represented as a triple (head, relation, tail). The \textit{head} refers to the persona entity, such as "a singer". The \textit{relation} defines the type of connection between the head and the tail, such as "Characteristic", "Routine or Habit", "Goal or Plan", "Experience", or "Relationship". The \textit{tail} provides the specific attribute or detail related to the head persona, like "good at singing" for a singer. This structured representation allows for the integration of persona knowledge into dialogue systems, improving their ability to generate more contextually relevant and engaging responses.

\subsection{Baseline Model}
A baseline model is provided, which is based on the DeBERTa model \citep{He2020DeBERTaDB} and finetuned on the ComFact dataset \citep{Gao2022ComFactAB}, utilizing a different knowledge graph than PeaCoK. The baseline model evaluates the relevance of both head and tail entities separately, outputting a positive label (true) only if both entities are relevant to the dialogue context. For details about ComFact, refer to Appendix Sec.~\ref{sec:comfact}.

\section{Methods}
We present a novel approach to persona knowledge linking in dialogues, leveraging LLMs for efficient data generation and knowledge distillation. Our method employs a sophisticated pipeline that harnesses the reasoning capabilities of LLMs to create a high-quality labeled dataset. This approach enables our student model to effectively distill knowledge from the emergent abilities of LLMs, resulting in a robust system capable of linking persona facts to conversations with high accuracy.

\subsection{SynCPKL Pipeline}
Given the absence of a pre-existing dataset for this task, we developed the SynCPKL Pipeline to create a high-quality dataset for training a classifier. Our approach leverages the PeaCoK dataset for commonsense persona fact, and PersonaChat \citep{Zhang2018PersonalizingDA} serves as the foundation for our conversation data \footnote{Previous efforts provided us with PersonaChat augmented with PeaCoK. The dataset can be found here: \url{https://github.com/Silin159/PeaCoK-PersonaChat?tab=readme-ov-file}}.

\subsubsection{Dataset Challenges and Quality Control}
Despite successfully identifying a suitable source dataset, several challenges emerged during our initial analysis. 
The dataset comprises dialogues between two interlocutors, each associated with a distinct set of persona attributes. However, we observed various inconsistencies and noise within the data. Use Table \ref{tab:persona_conversation} in the Appendix as an example. The dialogue suffers from irrelevant persona facts that are either not utilized or poorly integrated into the conversation. For Persona 1, two facts ("I wish I could live forever" and "I only date people taller than me") are completely unused. For Persona 2, the mention of loving iced tea feels forced and doesn't contribute meaningfully to the dialogue.

These issues could potentially compromise the efficacy of models trained on this data, particularly for tasks focused on persona-based dialogue understanding.

\subsubsection{LLM-based Data Generation and Labeling}
To address these challenges, we need a rigorous quality control process to create a dataset containing high-quality examples for accurately identifying the relevance of persona facts to a given conversation. Recent advancements in LLMs have demonstrated their potential as powerful labelers with human-like reasoning abilities \citep{Mitra2023Orca2T, Li2023TextbooksAA}. Leveraging this capability, we employed GPT-3.5-Turbo to generate synthetic data for our experiments.
Our data generation and labeling pipeline consists of the following key components:

\paragraph{Baseline Filtering to Prevent Imbalanced Data} 
Our initial approach to creating training data employed a naive heuristic: positive pairs were formed using the original corresponding persona facts, while negative pairs were constructed using the other participant's persona facts. However, this method was prone to incorrect labeling. For instance, two personas might share a common fact, which our heuristic would erroneously label as negative. Moreover, this approach led to an imbalanced dataset dominated by negative examples.

To address these limitations, we developed a baseline filtering model. We fine-tuned a DeBERTa model on the ComFact dataset. This baseline model was then used to predict relevance scores for each persona fact-conversation pair. To determine an appropriate threshold for creating soft labels, we utilized the online private test set provided by the shared task organizers, ensuring alignment with the task's objectives. This process resulted in a more balanced distribution of soft positive and negative labels before the official labeling.

\paragraph{Prompt Engineering and Iterative Refinement}
Our approach involved developing and evaluating multiple prompt templates, with Chain-of-Thought (CoT) prompting \citep{Wei2022ChainOT} proving most effective for GPT-3.5-Turbo. We iteratively processed and refined the dataset, starting with 10,000 examples and gradually expanding to 39,802 examples. This process included analyzing results, refining prompts based on observed patterns and errors, and continuously improving our prompting strategy to optimize data tagging performance.

\subsection{SynCPKL Dataset}
SynCPKL Pipeline demonstrates robust performance in curating a high-quality dataset, which we call SynCPKL. We generated two variants of the dataset:
(1) \textbf{SynCPKL$^{H}$}: Using the \textit{head} entity as the persona fact for GPT-3.5-Turbo to tag.
(2) \textbf{SynCPKL$^{T}$}: Using the \textit{tail} entity as the persona fact for GPT-3.5-Turbo to tag.

Each variant comprises 39,802 examples, with identical matching between the two. In the final version of \textbf{SynCPKL}, we combine both the relation head and tail as persona facts. An instance is labeled as true only when both the head and tail are independently verified as true.

\section{Experiments and Results}
We conducted a series of experiments to evaluate the performance of different models and input configurations on our persona-based knowledge-linking task. Our experiments utilized a subset of the private test set provided by the shared task organizers. Performance was measured using F1 score and accuracy (Acc).

\begin{table}[h]
\centering
\small
\begin{tabular}{@{}lllllll@{}}
\toprule
Version        & Dataset      & H      & T      & R  & F1    & Acc   \\ \midrule
Baseline       & ComFact      & $\checkmark$ & $\checkmark$ & $\times$     & 0.382 & 0.814 \\
$C^{H}$ & SynCPKL$^{H}$ & $\checkmark$ & $\times$     & $\times$     & 0.547 & 0.828 \\
$C^{t}$ & SynCPKL$^{T}$ & $\times$     & $\checkmark$ & $\times$     & 0.299 & 0.427 \\
$C^{H}$ $\land$ $C^{t}$       & -            & $\checkmark$ & $\checkmark$ & $\times$     & 0.548 & 0.849 \\
$C^{H,T}$           & SynCPKL      & $\checkmark$ & $\checkmark$ & $\times$     & 0.554 & 0.876 \\
$C^{R,H,T}$-NLI      & SynCPKL      & $\checkmark$ & $\checkmark$ & $\checkmark$ & 0.554 & 0.890 \\
$C^{R,H,T}$          & SynCPKL      & $\checkmark$ & $\checkmark$ & $\checkmark$ & \textbf{0.572} & 0.881 \\ \bottomrule
\end{tabular}
\caption{Different feature combinations of persona facts' results on the private test subset of the CPKL challenge. 
}
\label{exp:ablation test results}
\end{table}

\subsection{Experimental Setup}
We used DeBERTa-v3-large as our base model, with variations in the input representation of persona facts \footnote{The training was conducted on an NVIDIA RTX 3090 GPU with 24GB VRAM. The total training time on the full dataset was approximately 2 hours.}. We also included a comparison with a DeBERTa model fine-tuned for Natural Language Inference (NLI) \citep{laurer_building_2023}. 
The baseline model, trained on the ComFact dataset and based on DeBERTa-v3-large, was used for comparison.

\subsection{Feature Combination Analysis}

To investigate the impact of different input features on our model's performance, we conducted an ablation study examining various combinations of \textit{head}, \textit{relation}, and \textit{tail} entities with the following configurations:
\begin{itemize}
    \item Head only ($C^{H}$): A classifier trained on \textbf{SynCPKL$^{H}$}, using only the \textit{head} entity.
    \item Tail only ($C^{T}$): A classifier trained on \textbf{SynCPKL$^{T}$}, using only the \textit{tail} entity.
    \item Head Classifier ($C^{H}$) $\land$ Tail Classifier ($C^{T}$): Prediction is based on the \textsc{and} operation of models train on head and tail respectively.
    \item Head, Tail ($C^{H,T}$): A single classifier trained on \textbf{SynCPKL}, using both \textit{head} and \textit{tail} entities.
    \item Relation, Head, Tail ($C^{R,H,T}$): A classifier trained on \textbf{SynCPKL}, using the complete triple (\textit{head}, \textit{relation}, \textit{tail}) as the persona fact.
\end{itemize}

\subsection{Results \& Discussion}
Table \ref{exp:ablation test results} demonstrates that SynCPKL consistently outperforms Comfact across all metrics on the test subset, highlighting the efficacy of our synthetic data generation method.

Among the different feature combinations, the choice of Relation, Head, and Tail as persona fact yields the best overall performance (F1: 0.5729). This is intuitively straightforward that providing the model with complete information from KG is beneficial for accurate prediction. 

However, an intriguing pattern emerges when we examine the performance of partial input configurations. Contrary to expectations, the "Head-only" model ($C^{H}$'s F1: 0.547) performs comparably to the "Head, Tail" model ($C^{H, T}$'s F1: 0.554), both approaching the performance of the best configuration. This suggests that, in our test set, correct prediction of the head entity often leads to correct overall prediction. 

To investigate this phenomenon, we analyzed performance across different relation types (Table \ref{exp:results on relation}). We found that the "Head-only" model underperforms our best model for all relations except \textit{characteristic}. This exception is logical, as the \textit{characteristic} relation often allows for inference of the tail entity from the head entity, even when the tail is not explicitly mentioned in the conversation.

The poor performance of the "Tail-only" model (F1: 0.299) further emphasizes the importance of the head entity in our task. This asymmetry in the importance of head and tail entities warrants further investigation and may inform future model designs and data collection strategies.

Regarding the choice of pre-trained models, we found that fine-tuning from the original DeBERTa model is more effective than fine-tuning from the DeBERTa-NLI model. While the NLI task seems conceptually aligned with our task objective, we hypothesize that it may impose an overly strict definition of entailment, potentially misclassifying some instances of commonsense reasoning as false pairs.

\paragraph{Error Analysis.}

To gain deeper insights into the performance of our best-performing model, we conducted an error analysis. We randomly sampled 50 error cases from a total of 328 errors in our private test subset. This analysis revealed several key challenges for this task: 
1. Over-reliance on head mentions
2. Difficulty handling implicit and conditional information
3. Data quality issues

These findings highlight the need for improved integration of head and tail information, enhanced reasoning capabilities for implicit and conditional relationships, and more rigorous data curation processes to advance the model's performance in this complex task. See detailed error analysis in Appendix Sec.~\ref{sec:error}.

\section{Conclusion}
In this work, we introduce a novel approach that leverages Large Language Models (LLMs) for synthetic data generation and knowledge distillation in commonsense persona knowledge linking for dialogues. 
We present the first dataset, SynCPKL, in this task for PeaCok knowledge graph where SynCPKL is automatically constructed using our SynCPKL Pipeline. This pipeline offers a systematic and efficient method for utilizing LLMs in tasks lacking pre-existing datasets. Our comprehensive ablation studies reveal the most effective feature combinations and model configurations, showcasing superior reasoning capabilities for this complex task. Moreover, our final model, Deberta-SynCPKL, achieved first place in the CPKL challenge from the 6th Workshop on NLP for ConvAI.

\bibliography{anthology,custom}
\bibliographystyle{acl_natbib}

\appendix

\section{An Example Conversation with Persona Facts}
Table~\ref{tab:persona_conversation} shows a conversation between two people with their persona facts respectively. 

\section{ComFact: A Benchmark for Commonsense Fact Linking}
\label{sec:comfact}
ComFact is a benchmark for commonsense fact-linking in dialogues and storytelling. It addresses the challenge of identifying situationally relevant knowledge from a knowledge graphs, which is different from PeaCok. In ComFact, each data point consists of a context from a dialogue and a set of commonsense facts that need to be linked to this context. These facts are structured as triples (a head entity, a relation, and a tail entity).

\begin{table*}[h]
\centering
\begin{tabular}{|p{0.45\textwidth}|p{0.45\textwidth}|}
\hline
\multicolumn{2}{|l|}{\textbf{Persona Facts:}} \\
\hline
\textbf{Persona 1} & \textbf{Persona 2} \\
\hline
- I wish I could live forever. & - My mom is my best friend. \\
- I only date people taller than me. & - I have four sisters. \\
- I really like technology. & - I believe that mermaids are real. \\
- I like free diving. & - I love iced tea. \\
\hline
\multicolumn{2}{|l|}{\textbf{Conversation:}} \\
\hline
\multicolumn{2}{|p{0.9\textwidth}|}{
Persona1: Hi, how are you doing today?

Persona2: I am spending time with my 4 sisters what are you up to

Persona1: Wow, four sisters. Just watching game of thrones.

Persona2: That is a good show I watch that while drinking iced tea

Persona1: I agree. What do you do for a living?

Persona2: I'm a researcher I'm researching the fact that mermaids are real

Persona1: Interesting. I'm a website designer. Pretty much spend all my time on the computer.

Persona2: That's cool my mom does the same thing

Persona1: That's awesome. I have always had a love for technology.

Persona2: Tell me more about yourself

Persona1: I really enjoy free diving, how about you, have any hobbies?

Persona2: I enjoy hanging with my mother she's my best friend

Persona1: That's nice. Moms are pretty cool too.

Persona2: I'm also fascinated with mermaids
} \\
\hline
\end{tabular}
\caption{Example conversation between two personas}
\label{tab:persona_conversation}
\end{table*}

\section{Error Analysis}
\label{sec:error}
We conducted a comprehensive error analysis on our best-performing model to identify patterns in misclassifications and potential areas for improvement. This analysis involved manually reviewing a sample of incorrectly classified instances from the private test set.

We randomly sampled 50 error cases out of 328 total errors for our final model. Our observations are summarized as follows:

\begin{itemize}
    \item \textbf{Over-Reliance on \textit{Head} Mentions (10\%):} In 5 cases, the model incorrectly suggested relatedness when the \textit{head} was directly mentioned in the conversation, but the \textit{tail} was unrelated. This suggests that the model may be overly reliant on \textit{head} matches and fails to adequately consider \textit{tail} information. For example, an utterance "I am a relay racer" matched the \textit{head} "personx is a relay racer", but the model failed to recognize that the \textit{tail} was unrelated.
    
    \item \textbf{Implicit \textit{Head} Mentions with Conditional Tail Relevance (12\%):}  In 6 cases, the \textit{head} was vaguely implied within the conversation, and the \textit{tail} could only be inferred if the \textit{head} were true. This highlights the challenge of handling implicit information and conditional reasoning. For instance, with the \textit{head} "personx is a calligrapher" and \textit{tail} "writes in a beautiful script", the model struggled to make the connection that being a calligrapher implies writing beautifully.
    
    \item \textbf{Ambiguous Implications (4\%):} In 2 cases, both \textit{head} and \textit{tail} were very vaguely implied, presenting difficulties even for human annotators. This underscores the inherent ambiguity in some persona-based inferences and the potential for high disagreement in labeling.
    
    \item \textbf{Data quality issues (26\%):} 13 cases exhibited data quality problems, primarily where the gold reference suggested truth without supporting evidence in the conversation. For example, a conversation implying PersonX was a businessman was paired with a \textit{head} "personx is a rich investor who becomes rich" and \textit{tail} "made a large return on investment", neither of which were explicitly supported. This highlights the need for more rigorous data curation and annotation processes.
    
    \item \textbf{Model prediction errors (48\%):} 24 cases were clear model prediction errors, indicating room for improvement in the model's core reasoning capabilities.
\end{itemize}

Through the manual review, we identify a couple of insights:

\paragraph{Imbalance in Head-Tail Integration: Challenges in SynCPKL Labeling Strategy} The model demonstrates a tendency to over-rely on head matches, indicating a critical challenge in effectively integrating both head and tail information. This imbalance may stem from our current labeling strategy for SynCPKL, which designates a positive label only when both SynCPKL$^{H}$ and SynCPKL$^{T}$ are true. We hypothesize that the SynCPKL$^{T}$ component may lack sufficient informational content, potentially leading to false negatives in the LLM predictions.

\paragraph{Head-Tail Interdependence} The impact of different feature combinations exhibits significant variation across test subsets. 
This is particularly evident in the "Tail only" configuration, which demonstrates markedly poor performance on the test subset (F1: 0.2992). 
A plausible explanation for this lies in the inherent dependency of many tail entities on their corresponding head entities. For instance, consider a case where the \textit{head} is "a singer" and the \textit{tail} is "make music day to day." In such scenarios, the accuracy of the \textit{tail} classification is often contingent upon the correct identification of \textit{head} within the conversational context. 
This interdependence suggests that isolated \textit{tail} classification may be insufficient for robust performance, highlighting the importance of considering head-tail relationships in entity extraction tasks.
\paragraph{Data Quality Issues} Addressing issues in the training and evaluation data could lead to more reliable model performance and evaluation.
    
Our analysis indicates that while our model can distill some commonsense knowledge from LLMs, there are still significant opportunities to enhance its ability to effectively utilize LLM knowledge for this ambiguous task.

\begin{table}[]
\centering
\small
\begin{tabular}{@{}lllll@{}}
\toprule
               & \multicolumn{2}{l}{Head-only} & \multicolumn{2}{l}{Deberta-SynCPKL} \\ \midrule
Relation       & F1            & Acc           & F1               & Acc              \\ \midrule
characteristic & 0.5161        & 0.8370        & 0.4427           & 0.8678           \\
experience     & 0.5062        & 0.8357        & 0.5149           & 0.8994           \\
goal\_plan     & 0.4656        & 0.7815        & 0.5024           & 0.8592           \\
routine\_habit & 0.6430        & 0.8551        & 0.6892           & 0.8962           \\ \bottomrule
\end{tabular}
\caption{Results based on different relation in the private test subset}
\label{exp:results on relation}
\end{table}

\end{document}